\pdfoutput=1

\documentclass[11pt]{article}

\usepackage{acl}

\usepackage{graphicx}
\usepackage{amssymb}
\usepackage{multirow,tabularx}
\usepackage{changepage}

\definecolor{forestgreen}{HTML}{228B22}

\usepackage{times}
\usepackage{latexsym}
\usepackage{soul}

\usepackage[T1]{fontenc}

\usepackage[utf8]{inputenc}

\usepackage{microtype}

\usepackage{inconsolata}
\usepackage{url}
\usepackage{booktabs}
\usepackage{amsmath}
\usepackage{amssymb}
\usepackage{placeins}

\title{
DisentQA: Disentangling Parametric and Contextual Knowledge\\ with Counterfactual Question Answering}

\author{
Ella Neeman$^{1}$ \quad
Roee Aharoni$^{2}$ \quad
Or Honovich$^{3}$ \quad
Leshem Choshen$^{1}$ \quad \\
\textbf{Idan Szpektor}$^{2}$
\textbf{Omri Abend}$^{1}$ \quad \\\\
$^1$The Hebrew University of Jerusalem \quad
$^2$Google Research
\quad$^3$Tel Aviv University\\
{\tt \{ella.neeman, leshem.choshen, omri.abend\}@mail.huji.ac.il} \\
{\tt or.honovich@gmail.com} {\tt \{roeeaharoni,szpektor\}@google.com}
}

\begin{document}
\maketitle

\begin{abstract}


Question answering models commonly have access to two sources of ``knowledge'' during inference time: (1) \textit{parametric knowledge} -- the factual knowledge encoded in the model weights, and (2) \textit{contextual knowledge} -- external knowledge (e.g., a Wikipedia passage) given to the model to generate a grounded answer. Having these two sources of knowledge entangled together is a core issue for generative QA models as it is unclear whether the answer stems from the given non-parametric knowledge or not. This unclarity has implications on issues of trust, interpretability and factuality. In this work, we propose a new paradigm in which QA models are trained to disentangle the two sources of knowledge. Using counterfactual data augmentation, we introduce a model that predicts two answers for a given question: one based on given contextual knowledge and one based on parametric knowledge. Our experiments on the Natural Questions dataset show that this approach improves the performance of QA models by making them more robust to knowledge conflicts between the two knowledge sources, while generating useful disentangled answers.

\end{abstract}

\section{Introduction}

\begin{figure}[!ht]
\begin{center}
\small
\resizebox{0.9\columnwidth}{!}{%
\begin{tabular}{l}
\multicolumn{1}{c}{\textbf{Question:}  Who is the guy on Keeping Up with}  \\ 
\multicolumn{1}{c}{the Kardashians?}  \\ 
\\
\multicolumn{1}{c}{\textbf{Factual}}  
\\ \hline
\multicolumn{1}{|l|}{\begin{tabular}[c]{@{}l@{}}
\textbf{Context:} \textcolor{blue}{\textbf{Jonathan Cheban}}  (born c. 1974)\\
is a reality - television star and entrepreneur. \\
He is noted for his recurring role on the show \\
Keeping Up with the Kardashians and its \\
spinoffs.
\end{tabular}}  \\
\multicolumn{1}{|l|}{\textbf{Contextual Answer:} \textcolor{blue}{\textbf{Jonathan Cheban}} } \\
\multicolumn{1}{|l|}{\textbf{Parametric Answer:} \textcolor{forestgreen}{\textbf{Scott Disick}}} \\ \hline
\\
\multicolumn{1}{c}{\textbf{Counterfactual}}  
\\ \hline
\multicolumn{1}{|l|}{\begin{tabular}[c]{@{}l@{}}
\textbf{Context:} \textcolor{red}{\textbf{Jason Momoa}} (born c. 1974)\\
is a reality - television star and entrepreneur.\\
He is noted for his recurring role on the show \\
Keeping Up with the Kardashians and its \\
spinoffs.\end{tabular}} \\
\multicolumn{1}{|l|}{\textbf{Contextual Answer:} \textcolor{red}{\textbf{\textbf{Jason Momoa}}}}\\ 
\multicolumn{1}{|l|}{\textbf{Parametric Answer:} \textcolor{forestgreen}{\textbf{Kanye West}}}
\\ \hline
\end{tabular}
}
\end{center}
\caption{Example outputs from our disentangled QA model on the Natural Questions dataset. The model generates two answers at once -- one based on the given context (blue and red), and another based on its parametric knowledge (green). Jonathan Cheban, Scott Disick and Kanye West are all prominent male characters on the show, while Jason Momoa never appeared in it.
}
\label{fig:intro_example}
\vspace{-15px}
\end{figure}

 Question answering (QA) systems are important in many real-world scenarios that require quick access to large bodies of knowledge like the web.
Much of the recent progress on QA stems from using pretrained models, shown to implicitly store knowledge in their parameters \citep{roberts-etal-2020-much}.

As a result, QA models have access to two knowledge sources when generating an answer: (1) \textit{parametric knowledge} -- knowledge encoded (or ``memorized'') in the model parameters, and (2) \textit{contextual knowledge} -- knowledge encapsulated within external textual sources given to the model at inference time as the context of the question, such as paragraphs retrieved based on the question.

Disentangling the knowledge sources allows detecting and handling \textit{knowledge conflicts}.
Without disentanglement the behaviour when the contextual and parametric answers contradict each other is undefined and often erroneous. Unfortunately, both answers may be wrong at times, resulting in system errors.
More issues arise with lower quality context retrieval \citep{longpre-etal-2021-entity} and the parametric knowledge may fail when the answer changes over time \citep{dhingra-etal-2022-time}. For example, ``who is the president of the US?'', may result in knowledge conflicts if the parametric knowledge is stale. 

Another related issue is \textit{answerability}, where a model generates an answer despite no answer being present in the contextual knowledge, resulting in ungrounded answers \cite{rajpurkar-etal-2018-know,asai-choi-2021-challenges, sulem-etal-2021-know-dont, kim-etal-2021-linguist}, i.e., answers that are not attributable to the given source \cite{rashkin2021measuring}. 
All the above issues and the inability to know whether an answer was generated based on contextual knowledge or the parametric knowledge, give rise to issues of user trust -- especially as models are prone to mimicking human falsehoods \cite{lin-etal-2022-truthfulqa}.





In this work, we propose a new paradigm for generative QA models that alleviates the above issues by encouraging \textit{disentanglement} of parametric knowledge from contextual knowledge. Specifically, we propose a single model that generates two answers to a given question -- a parametric answer and a contextual answer, in one-fell-swoop. Figure \ref{fig:intro_example} exemplifies this.
To achieve this, we use two training data augmentation methods: (1) Counterfactual Data Augmentation \cite{longpre-etal-2021-entity}, obtained by automatically altering facts in a given QA corpus to decrease reliance on parametric knowledge, and (2) Answerability Augmentation, where we train the model to abstain from answering when no answer is present in the contextual knowledge.


We perform a thorough analysis of our proposed approach while controlling for different training conditions and model size.
Our experiments on the Natural Questions dataset \cite{kwiatkowski-etal-2019-natural} show that disentangled models are able to provide different answers to the same question -- contextual answers based on the external contextual knowledge, but also different-but-useful parametric answers based on their vast parametric knowledge acquired during pre-training. 
In addition, we found disentangled models to have better performance w.r.t. knowledge conflicts than vanilla models. We report limitations of the work in App.~\ref{sec:limitations}.
We hope this work will encourage more progress on disentangling knowledge sources in QA and NLP in general, towards more faithful and useful applications.\footnote{Our code and data are publicly available at \\\url{https://github.com/ellaneeman/disent_qa}} 

\section{Separating Parametric Knowledge from Contextual Knowledge}
\label{sec:data}

We next describe our methodology for disentangling parametric knowledge\footnote{We acknowledge that using the term ``knowledge'' when discussing a neural network that predicts tokens may be anthropomorphic. However, we find this abstraction useful, and it is common in  recent literature \citep{petroni-etal-2021-kilt}.} from contextual knowledge in generative QA models. 
We first introduce the overall approach, and then describe our augmentation of a typical QA training set to support this approach.

\subsection{Predicting Disentangled Answers}


We are interested in exploring whether a single model can predict two types of answers in a single output: one based on the contextual knowledge, followed by one based on the parametric knowledge. If this succeeds, we can say that the model has disentangled the two knowledge sources, possibly improving its performance by alleviating issues like knowledge conflicts or hallucinated answers. This disentanglement is also useful for explaining and debugging the model's answers, and for improving user trust in the provided answers, e.g., by reporting agreement or conflict: ``\emph{According to this external source, the answer is} A. \emph{According to my parametric knowledge, the answer is} B''.


To enable this capability, we create a QA training set with examples consisting of a question and a context paragraph as input and two answers -- a parametric answer and a contextual answer -- as output. To this end, we start with a standard QA training set, where we assume that at least for some of the questions, the knowledge needed for predicting the correct answer was obtained during pre-training of the language model that we fine tune for the task.
We then create three types of training examples from the original QA examples. In all these example types, the parametric answer is the original answer to the question (as it appeared in the original training data), and they differ only in their input context and therefore in their contextual answers: (1) \textbf{Factual Examples} -- the context and contextual answers are taken from a QA dataset as-is. (2) \textbf{Counterfactual Examples} (Section~\ref{sec:counterfactual}) -- the context is altered to induce a new (counterfactual) answer. (3) \textbf{Unanswerable Examples} (Section~\ref{sec:answerability}) -- the model is trained to abstain from answering a contextual answer
when given one of two types of contexts: empty or random.

\begin{table}
\centering
\small
\resizebox{0.98\columnwidth}{!}{%
\begin{tabular}[th!]{l|l|l}
\textbf{Example Type} & \textbf{Input Context} & \textbf{Contextual Answer} \\
\hline
factual & original context & original answer \\
counterfactual & counterfactual & counterfactual answer \\
empty & empty & \textit{unanswerable} \\
random  & random & \textit{unanswerable} \\ 
\end{tabular}
}
\caption{Example types for training a QA model to provide both parametric and contextual answers.}
\label{tab:scenarios}
\vspace{-10pt}
\end{table}

Table~\ref{tab:scenarios} summarizes our training example types and their differences and Figure \ref{fig:factual_and_random_examples} presents concrete examples. We hypothesize that training a QA model on all of these example types would encourage it to disentangle the representation of its two knowledge sources and generate different answers to the same question when there's a mismatch between the two sources. 



\subsection{Counterfactual Data Augmentation}
\label{sec:counterfactual}

To generate counterfactual examples where the parametric answer differs from the contextual answer, we adopt the counterfactual data augmentation framework of \citet{longpre-etal-2021-entity} which was proposed to mitigate knowledge conflicts in QA models. 
There, for each example -- a (question, context, answer) triplet, a counterfactual example is created by replacing the answer instances in the context with a different answer (which does not appear in the original context). The new answer is used as the contextual answer, training the model to predict the new answer for this context without changing the question. For example in Figure \ref{fig:factual_and_random_examples}, ``Ukraine'' was replaced with ``Brazil''. 

\subsection{Answerability Augmentation}
\label{sec:answerability}

Counterfactual examples do not address cases where the model should abstain from answering when no relevant answer is present in the input context. We hypothesize that improving the ability of models to abstain from answering when given irrelevant context should further encourage the disentanglement of parametric and contextual knowledge, as they should steer away from generating hallucinated contextual answers based on the parametric knowledge, while still exposing relevant parametric knowledge via the parametric answer.

Several previous works focused on this \textit{answerability} aspect in QA \citep{sulem-etal-2021-know-dont, kim-etal-2021-linguist}, with \citet{asai-choi-2021-challenges} showing that when models are provided with a gold retrieved paragraph and the ability to decline answering, they outperform human annotators. Following this line of work, and similarly to SQuAD 2.0 \citep{rajpurkar-etal-2018-know} in the extractive QA setting, we create additional training examples for which the model should explicitly predict that no answer is present in the external source. We replace the original context in training examples with either an empty context or with a randomly sampled context, which is not expected to include information useful to generate the original answer, as shown in Figure \ref{fig:factual_and_random_examples}.

\begin{figure}[t]
\begin{center}
\small
\begin{tabular}{p{0.95\columnwidth}}
\multicolumn{1}{c}{\textbf{Question:}  What country shares borders with}  \\ 
\multicolumn{1}{c}{both Belarus and Romania?}  \\ 
\multicolumn{1}{c}{\textbf{Factual}}  
\\ \hline
\multicolumn{1}{|l|}{\begin{tabular}[c]{@{}p{0.95\columnwidth}@{}}
\textbf{Context:}  \textcolor{blue}{\textbf{Ukraine}} borders with seven countries:
Poland, Slovakia, Hungary, Romania, Moldova, 
Russia and Belarus.
...
\end{tabular}}  \\
\multicolumn{1}{|l|}{\textbf{Contextual Answer:} \textcolor{blue}{\textbf{Ukraine}} } \\
\multicolumn{1}{|l|}{\textbf{Parametric Answer:} \textcolor{forestgreen}{\textbf{Ukraine}}} \\ 
\hline

\multicolumn{1}{c}{\textbf{Counterfactual}}  
\\
\hline
\multicolumn{1}{|l|}{\begin{tabular}[c]{@{}p{0.95\columnwidth}@{}}
\textbf{Context:}  \textcolor{blue}{\textbf{Brazil}} borders with seven countries:
Poland, Slovakia, Hungary, Romania, Moldova, 
Russia and Belarus.
...
\end{tabular}}  \\
\multicolumn{1}{|l|}{\textbf{Contextual Answer:} \textcolor{blue}{\textbf{Brazil}} } \\
\multicolumn{1}{|l|}{\textbf{Parametric Answer:} \textcolor{forestgreen}{\textbf{Ukraine}}} \\ 
\hline

\multicolumn{1}{c}{\textbf{Empty}}  
\\ \hline
\multicolumn{1}{|l|}{\begin{tabular}[c]{@{}p{0.95\columnwidth}@{}}
\textbf{Context:} 
\end{tabular}} \\
\multicolumn{1}{|l|}{\textbf{Contextual Answer:} \textcolor{gray}{\textbf{\textbf{Unanswerable}}}}\\ 
\multicolumn{1}{|l|}{\textbf{Parametric Answer:} \textcolor{forestgreen}{\textbf{Ukraine}}}                 
\\ 
\hline

\multicolumn{1}{c}{\textbf{Random}}  
\\ \hline
\multicolumn{1}{|l|}{\begin{tabular}[c]{@{}p{0.95\columnwidth}@{}}
\textbf{Context:} The epic, traditionally ascribed to the Hindu
sage Valmiki, narrates the life of Rama, the
legendary prince of 
...
\end{tabular}} \\
\multicolumn{1}{|l|}{\textbf{Contextual Answer:} \textcolor{gray}{\textbf{\textbf{Unanswerable}}}}\\ 
\multicolumn{1}{|l|}{\textbf{Parametric Answer:} \textcolor{forestgreen}{\textbf{Ukraine}}}                 
\\ 
\hline
\\
\end{tabular}
\end{center}
\vspace{-15px}
\caption{Training examples derived from a single Natural Questions example. The top example is the original, requiring the contextual and parametric answers to be identical. The second is a counterfactual example generated by altering Ukraine to Brazil. The bottom two replace the context to be random or empty, and accordingly the contextual answer to be \emph{unanswerable}.
}
\label{fig:factual_and_random_examples}
\vspace{-15px}
\end{figure}

\section{Experimental Setup}







\subsection{Natural Questions}

We base our experiments on the
Natural Questions \citep[NQ;][]{kwiatkowski-etal-2019-natural} dataset. NQ is a dataset compiled from questions naturally queried by users of the Google search engine, hence used to test the real-world applicability of QA models. Each example includes a question, a passage (``long answer''), and a short answer that can be inferred from the passage. NQ enables benchmarking QA systems that include a retrieval component to obtain relevant passages from a knowledge-base given a question. We focus on the QA model itself and not on the retrieval model, so we always use the ``gold'' passage as the context, assuming an oracle retrieval system. We use the examples that have both a gold passage and a short answer (35\% of the data). We use an example if at least one out of the five annotators found the corresponding passage suitable to answer the question.

\begin{table}[ht!]
\small
\resizebox{0.98\columnwidth}{!}{%


\begin{tabular}{|l|l|l|l|l|}
\hline
 & Factual & Counterfactual & Empty & Random \\ \hline
Train & 85,540 & 30,653 & 85,540 & 85,540 \\ \hline
Validation & 21,386 & 7,698 & 21,386 & 21,386 \\ \hline
Test & 1,365 & 1,365 & 1,365 & 1,365 \\ \hline
\end{tabular}}
\caption{Dataset size (columns) per split (rows). 
}
\label{tab:data}
\vspace{-10px}
\end{table}


\subsection{Counterfactual Example Generation}

To create counterfactual examples we follow the substitution framework proposed in \citet{longpre-etal-2021-entity} which generates counterfactual examples given a QA dataset. It modifies the context to alter the answer. This process includes (1) identifying named entity answers, and (2) replacing all appearances of the answer in the context by a substituted entity. We use the ``corpus-substitution'' policy \citep{longpre-etal-2021-entity}, which replaces answers with other answers of the same entity type, sampled from the same corpus. This process resulted in 30,653 counterfactual examples for training, and additional 7,698 examples for validation induced from the NQ training data. The same process is done on NQ dev set, producing 1,365 altered examples. Table \ref{tab:data} details the full statistics of our induced dataset. 
We note that all additional examples are based on a subset of questions already appearing in the training/dev sets, so no new questions are introduced in this process. For a fair comparison between the 4 datasets, we keep in the test set just the examples that induced the counterfactual dataset.


\subsection{Metrics and Evaluation}

We evaluate our model on the NQ development set using Exact Match (accuracy) \citep{rajpurkar-etal-2016-squad}. We report the following metrics:

\begin{enumerate}
\item \emph{Contextual Answer Quality}: Accuracy on the original NQ dev set. We compare the contextual answer to the expected (original) answer.

\item \emph{Robustness} (to knowledge conflicts): the accuracy of the contextual answer when evaluated on counterfactual data (altered examples from NQ dev).  We compare the contextual answer to the expected (altered) answer.

\item \emph{Answerability}: the accuracy of the model in abstaining from giving a contextual answer when given a random or empty context. Defined as the as accuracy for predicting the special token ``unanswerable'' on such examples.

\item \emph{Answer Separation}: The extent of the disentanglement -- percentage of cases where the parametric answer is different from the contextual answer

\item \emph{Parametric Answer Quality}: accuracy of the parametric answers on the NQ dev set.

\end{enumerate}

\begin{table*}
\begin{small}
\begin{center}
\begin{tabular}[t]{|l|l|l|l|ll|}

\hline
 & \textbf{Model Name} & \textbf{Output Format} & \textbf{Training Data} & \textbf{Contextual} & \textbf{Parametric} \\ \hline\hline
(s) cb & closed-book & \multirow{2}{*}{baselines} & empty & - & \checkmark \\ \cline{1-2} \cline{4-6} 
(s) f & single answer, factual &  & factual & \checkmark & - \\ \hline\hline
(s) f+cf & + counterfactual & \multirow{3}{*}{single answer} & factual, counterfactual & \checkmark & - \\ \cline{1-2} \cline{4-6} 
(s) f+a & + answerabilty &  & factual, empty, random & \checkmark & - \\ \cline{1-2} \cline{4-6} 
(s) f+cf+a & + counterfactual + answerabilty &  & all & \checkmark & - \\ \hline\hline
(m) f+cf & + counterfactual & \multirow{3}{*}{multi answer} & factual, counterfactual & \checkmark & \checkmark \\ \cline{1-2} \cline{4-6} 
(m) f+a & + answerabilty &  & factual, empty, random & \checkmark & \checkmark \\ \cline{1-2} \cline{4-6} 
(m) f+cf+a & + counterfactual + answerabilty &  & all & \checkmark & \checkmark \\ \hline

\end{tabular}
\caption{Baselines and models described by their training data and output format. Specifically, the models differ by the context types they see during training, denoted by acronyms separated by the "+" sign, and the number of answers they are required to predict (single/multi answer), denoted by (s) or (m).  }
\label{tab:baselines}
\end{center}
\end{small}
\vspace{-15px}
\end{table*}

\subsection{Models}
The QA models listed in Table~\ref{tab:baselines} were trained on the example types described in Section \ref{sec:data} -- either on all of them or some of them for ablation.
We encode the question and context as the input sequence and decode the answer(s) as the output sequence. We fine-tune T5 models \cite{2020t5} of two sizes (
Large -- 770M parameters, XXL -- 11B parameters), as we found that model size greatly affects the amount of parametric knowledge available to the model. More details about the models are available in App.~\ref{ap:sec:setup}.
We train the following models: 


\paragraph{Closed-Book Baseline.}
 A closed-book (cb) model that given a question and an empty context predicts a \textit{single} answer. The model has no access to external knowledge and it relies only on the knowledge encoded in its parameters to generate an answer. This baseline measures the relevance of the parametric knowledge to the tested questions \citep{roberts-etal-2020-much}.

\paragraph{Single, Factual (Vanilla) Baseline.}
The standard contextual setting: given a question and a context passage, the model predicts a \emph{single} answer. This model is trained only on \emph{factual} examples.


\paragraph{Single, Factual + Counterfactual.}
A contextual model that predicts a \textit{single} answer given the question and the context. On top of the \emph{factual} examples that the Vanilla model is trained on, this model is also trained on \emph{counterfactual} examples.

\paragraph{Single, Factual + Answerabilty.}
A contextual model that predicts a \emph{single} answer given the question/context input. On top of the \emph{factual} examples, this model is trained on \emph{empty} and \emph{random} context examples to learn to abstain from answering.

\paragraph{Single, Factual + Counterfactual + Answerabilty.}
A contextual model that predicts a \emph{single} answer given the the question/context input. On top of the \emph{factual} examples, this model is trained on all the training-data augmentation examples: \emph{counterfactual}, \emph{empty} and \emph{random} context.


\paragraph{Multi, Factual + Counterfactual.}
A contextual model that predicts \emph{two answers} given the question and the context, in the format of ``\emph{contextual: <contextual answer>, parametric: <parametric answer>}''. The model is trained on \emph{factual} and \emph{counterfactual} examples to predict the first answer based on the context and the second answer from the parametric knowledge (see Table~\ref{tab:scenarios}).

\paragraph{Multi, Factual + Answerabilty.}
A contextual model that predicts \emph{two answers} given the question and the context, in the format described above. The model is trained on \emph{factual} examples and \emph{empty} and \emph{random} context examples, to learn to abstain from offering a contextual answer in such cases. 

\paragraph{Multi, Factual + Counterfactual + Answerabilty.}
A contextual model that predicts \emph{two answers} given the question and the context, in the above format. It is trained on the \emph{factual}, \emph{counterfactual}, \emph{empty} and \emph{random} context examples, as described in Table~\ref{tab:scenarios}.

\section{Results}
We report and discuss here the results for the T5-11B models. The T5-Large models' results are presented in Appendix \ref{ap:sec:results}.
\subsection{Contextual Answer Quality}

We evaluate how the proposed changes affect the standard NQ settings by evaluating the contextual answers on the factual (unchanged) test set. As shown in Table~\ref{tab:results_contextual} on the ``factual'' column, all models maintain the ability to give correct answers based on the context, with accuracy ranging between 78.1 to 80.81. Adding answerability seems to slightly degrade performance, while adding this important capability. 
Counterfactual augmentation (the ``(s) f+cf'' model) presents improvements over the vanilla model, in accordance with the findings of \citet{longpre-etal-2021-entity}.
Adding the parametric answer (``(s)'' vs. ``(m)'' models) has little effect on the results, while again adding a new capability.

\subsection{Robustness} \label{subsec:results_robustness}

We measure model robustness to knowledge conflicts when given counterfactual examples, where it should adhere to the altered context. As Table~\ref{tab:results_contextual} shows on the ``counterfactual'' column, the vanilla model performs worst. This may indicate model confusion caused by conflicting parametric and contextual knowledge. Counterfactual augmentation improves performance in this setting, and adding answerability boosts performance even further by  5.35 points, resulting in a score of 84.98. Predicting the parametric answer does not seem to help in this setting but also does no harm when used together with the data augmentation methods. We conclude that adding both answerabitliy and counterfactual augmentation improves the model robustness, and their effect is complementary.

\subsection{Answerability}\label{subsec:results_answerability}

We measure \emph{Answerabilty}, defined as the accuracy score for predicting the special token ``unanswerable" in the contextual answer, in 
 Table~\ref{tab:results_answerabilty}. When given an empty context, all models correctly predict ``unanswerable'' in more than 99\% of the cases. Random, irrelevant context is more challenging -- only models trained with counterfactual data (``f+cf+a'') achieve high accuracy, and others (``f+a'') only achieve 27.69 and 35.6 accuracy, again showing how the augmentation methods are complementary.

\begin{table}[ht]
\centering
\begin{small}
\begin{tabular}{|l|c|c|}
\hline
& Factual $\uparrow$ & Counterfactual $\uparrow$ \\ \hline
(s) f (vanilla) & 79.34 & 66.81 \\\hline \hline
(s) f+cf & 80.73 & 79.63 \\ \hline
(s) f+a      &   80.81 &          69.30 \\ \hline
(s) f+cf+a & 78.32 & 84.98 \\ \hline
 \hline
(m) f+cf & 80.37 & 76.92 \\ \hline
(m) f+a & 80.22 & 64.62 \\ \hline
(m) f+cf+a & 78.10 & 84.91 \\ \hline
\end{tabular}
\end{small}

\caption{Accuracy (in percent) of the contextual answers on the factual and counterfactual datasets.}
\label{tab:results_contextual}
\vspace{-15pt}
\end{table}

\begin{table}[ht]
\centering
\begin{small}
\begin{tabular}{|l|c|c|}
\hline
 & Empty $\uparrow$ & Random $\uparrow$ \\ \hline
 (s) f+a      & 100.00 &  27.69 \\  \hline
(s) f+cf+a & 100.00 & 99.34 \\ \hline\hline
(m) f+a & 100.00 & 35.60 \\ \hline
(m) f+cf+a & 100.00 & 99.49 \\ \hline
\end{tabular}
\end{small}
\caption{Accuracy for predicting the special token ``unanswerable" in the contextual answer.
}
\label{tab:results_answerabilty}
\vspace{-10px}
\end{table}

\begin{table}[ht]
\centering
\resizebox{0.98\columnwidth}{!}{%
\begin{tabular}{|l|r|r|r|r|}
\hline
 & Factual $\uparrow$ &  Counterfactual $\downarrow$ & Empty $\downarrow$ & Random $\downarrow$ \\ \hline
(m) f+cf & 99.93 & 92.45 & 99.93 & 99.71 \\ \hline
(m) f+a & 99.85 & 99.71 & 0 & 64.32 \\ \hline
(m) f+cf+a & 93.55 & 18.46 & 0 & 0.29 \\ \hline
\end{tabular}}
\caption{Answer Separation: similarity between the contextual and parametric answer (percentage of cases where the two answers are identical). 
}
\label{tab:results_similarity}
\vspace{-15pt}
\end{table}

\subsection{Answer Separation}


We report \emph{Answer Separation} which is the similarity between the contextual and the parametric answers on a given test set. On the counterfactual test set, contextual and parametric answers should differ -- so lower ($\downarrow$) similarity is better, while on the factual test set the two should coincide, so higher ($\uparrow$) similarity is expected. The results in Table~\ref{tab:results_similarity} demonstrate that the ``(m) f+cf+a'' model successfully performs disentanglement: the contextual and parametric answers largely differ on the counterfactual data, with an average similarity of 18.46\%. Other models fail to disentangle the contextual and parametric knowledge, showing again that all of the suggested augmentations are essential and complementary for disentanglement. 
On the factual test set, parametric and contextual answers are mostly identical (with more than 99\% similarity), as expected.
 In both empty and random context scenarios, the contextual answer should be ``unanswerable'', while the parametric answer should be derived from memorized knowledge. Unsurprisingly, the model that is not trained for answerability -- ``(m) f+cf'' -- wrongly predicts identical contextual and parametric answers in those cases, with similarity higher than 99. For the two other models, ``(m) f+a'' and ``(m) f+cf+a'' results are consistent with those observed in section \ref{subsec:results_answerability}, where the full augmentation is best, and random contexts are more challenging.

\begin{table}[ht]
\resizebox{\columnwidth}{!}{%
\centering
\begin{tabular}{|l|c|c|c|c|}
\hline
 & Factual $?$ &  Counterfactual $\uparrow$ & Empty $\uparrow$ & Random $\uparrow$ \\ \hline
(s) cb & - & - & 27.69& - \\ \hline
(m) f+cf  &   80.37 &           9.23 & 20.73 &  13.92 \\\hline
(m) f+a   &   80.22 &           5.93 & 25.35 &  23.15 \\\hline
(m) f+cf+a &   74.87 &          44.69 & 31.14 &  30.18 \\\hline
\end{tabular}}
\caption{Accuracy (in percent) of parametric answers.}
\label{tab:results_parametric_11b}
\vspace{-15px}
\end{table}

\subsection{Parametric Answer Quality}
\label{subsec:results_parametric}

We evaluate the ability of the models to answer based on their parameters when given an empty context, comparing the parametric answer to the original answer on NQ. We evaluate all models that can predict a parametric answer (\checkmark in Table~\ref{tab:baselines}). Results are shown in Table~\ref{tab:results_parametric_11b}, in the ``empty'' column.

The baseline in this setting is the ``(s) cb'' model, whose accuracy is 27.69. While it is not clear why a model that was trained to use both contextual and parametric knowledge should perform better in this setting, the ``(m) f+cf+a'' improves over the baseline in 3.5 points. 
We would expect a model to score the same on all example types, because the model here should generate an answer that comes from the parameters, irrespective of the context. However, we find that parametric answers still change with the provided context; for random context, the results are slightly lower than the ones with an empty context in all models. With counterfactual context the results are lower for models without answerability, but higher when introducing all augmentation methods together, possibly showing that the model learns to use ``hints'' from the counterfactual context. Finally, when given the factual context, the parametric answer quality is much higher as it is trained to imitate the contextual answer in this scenario. Interestingly, in the model that uses all augmentation methods, this imitation happens less often, which may point to better disentanglement (hence the ``?'' in the ``factual'' column title, as better is not necessarily about higher accuracy, but rather about different answers).


\begin{table*}
\centering
\begin{small}
\begin{tabular}{|l|c|c|c|c|}
\hline
 & Factual (diff) & Counterfactual (diff) &  Empty (diff) & Random (diff) \\\hline

          (s) cb &   - & - &  9.76  (17.93) &  - \\\hline
         (m) f+cf &  77.51 (2.86) &          2.07 (7.16) &  7.40 (13.33) &  5.03 (8.89) \\\hline
          (m) f+a &  78.99 (1.23) &          1.48 (4.45) & 10.06  (15.29) &  8.58 (14.57) \\\hline
        (m) f+cf+a &  68.05 (6.82) &         12.72 (31.97) &  7.40  (23.74) &  7.10 (23.08) \\\hline

\end{tabular}
\end{small}
\caption{Parametric Answer accuracy predicted on the No Answer Overlap (NAO) dev set. In brackets, difference from total accuracy reported on the Dev set (Answer overlap + No Answer Overlap).}
\label{tab:nao_param}
\vspace{-15pt}
\end{table*}

\section{Analysis}
\subsection{Answer Overlap in NQ}
\label{subsec:results_nao}

Different questions that have identical answers in the training and test data may create unwanted artifacts. We follow \citet{lewis-etal-2021-question} and split the test sets into Answer Overlap (AO) / No Answer Overlap (NAO) subsets, that contain only reference answers that appear/do not appear in the training set, and recompute our metrics on the more challenging NAO subset.

We find that 
\emph{Contextual Answer Quality} and \emph{Robustness} present similar trends, but all models perform slightly worse on the factual NAO dataset in comparison to the AO+NAO full factual dataset. In the counterfactual NAO dataset, the models perform slightly better when we ignore AO examples. That might indicate that, when available, the model uses some memorized knowledge in its contextual prediction. See Appendix \ref{ap:sec:results} for the full results.


For \emph{Parametric Answer Quality} we see differences on the NAO datasets. Table~\ref{tab:nao_param} shows that for the counterfactual, empty and random contexts, the differences in accuracy between the NAO subset and the entire dataset are significant. This suggests that when models successfully predict the expected parametric answer with random or empty context, many times this is due to answer overlap between the training and the test data (but not always, as the numbers are non-zero in all cases).

\subsection{Effect of Model Size}


We replicate our experiments with T5-Large (App.~\ref{ap:sec:results}), and find that the T5-11B models perform better in all cases, and that the trends hold for the different model variations. 

\begin{table*}[t]
\small
\resizebox{0.98\textwidth}{!}{%

\begin{tabular}{|p{0.014\textwidth} |p{0.4\textwidth} | p{0.18\textwidth}| p{0.15\textwidth}|| p{0.15\textwidth}|}
\hline
 & Context & Question & Contextual Answer & Parametric Answer \\ \hline
1 & A number of \st{Michelangelo} \textit{John Locke}'s works of painting, sculpture and architecture rank among the most famous in existence…He sculpted…the Pietà and David…he also created...scenes from Genesis on the ceiling of the Sistine Chapel in Rome… & Who created the pieta and also painted the ceiling of the Sistine chapel? & John Locke & Michelangelo  \\ \hline
2 & Mission commander ...\st{Neil Armstrong} \textit{Freddie Highmore} became the first human to step onto the lunar surface... & Who took the first steps on the moon in 1969?& Freddie Highmore &  Neil Armstrong \\ \hline
3 & Psychoanalysis...was established in the early 1890s by Austrian neurologist \st{Sigmund Freud} \textit{Robert Remak}... & Who is regarded as the founder of psychoanalysis? & Austrian neurologist Robert Remak &  \textit{Austrian neurologist Robert Remak} \\ \hline
4 & Table conveying: \st{Johnny Depp} \textit{Ben Savage} starred in Pirates of the Caribbean & Who starred in the Pirates of the Caribbean? & \textit{Johnny Depp} &  Johnny Depp \\ \hline

\end{tabular}
}
\caption{Example answers of (m) f+cf+a.  Contexts are taken from the counterfactual examples. Replaced words are \st{striked through} and replacements  and wrong answers are \textit{italicized}.}
\label{tab:examples_con-par}
\vspace{-10px}
\end{table*}

\subsection{Manual Analysis}
\paragraph{Disentanglement.}
To get a better impression of how disentanglement works, we show some examples of parametric vs. contextual answers in Table~\ref{tab:examples_con-par}. Often, ``(m) f+cf+a'' is robust to knowledge conflicts, and can disentangle the two sources of knowledge -- contextual and parametric (Ex. 1-2). However, sometimes knowledge leaks from the contextual to the parametric knowledge (Ex. 3) or the other way around (Ex. 4).

\paragraph{Error Analysis.}
First, we examine the performance decrease of the ``(m) f+cf+a'' model on the factual data relative to vanilla (\S\ref{tab:results_contextual}). We analyze the 73 examples in which the model failed on the factual data while the vanilla model succeeded. In 14 of these examples, the model received a 0 score despite being correct (e.g., answering ``Napoleon'' when the reference was ``Napoleon Bonaparte''). 8 errors  were introduced due to the addition of answerability, where the model predicted ``unanswerable'' when an answer was in fact present in the context. In 12 cases, the wrong prediction is not part of the context. We observed 6 cases where there was more than one correct answer, and the model did not select the expected one. For example, given the question ``\textit{Who wrote the song photograph by Ringo Starr?}'' and the context: \textit{``Photograph is a song by English musician Ringo Starr... Starr co-wrote the song with George Harrison...}'', the model selected the valid answer ``George Harrison'', but the expected answer was ``Ringo Starr''. The remaining 33 examples are wrong answers, taken from the context. Half of them are challenging cases where the context is a table, the expected answer contains numbers, or the question is unclear.

Next, we look into the gap between the ``(m) f+a'' model and the ``(m) f+cf+a'' model in detecting unanswerable cases, when provided with random context (\S\ref{tab:results_contextual}). While ``(m) f+cf+a'' easily detects such cases, ``(m) f+a'' fails in 64.4\% of them, despite being trained on random contexts. This shows that the augmentation methods are complementary, as only the ``(m) f+cf+a'' succeeded to detect the cases. When failing to predict ``unanswerable'', we observe that the model invariably predicts the same contextual and parametric answers. We thus conclude that ``(m) f+a'' did not learn to perform disentanglement, and instead copied the parametric answer to the contextual answer in many cases.`` 

For example, given ``Who won the 2018 women's Royal Rumble match?'', the correct parametric answer is ``Asuka'', while the model answered ``Naomi'' in both answers (Naomi is a professional wrestler who participated in the contest). 

In 176 out of 879 wrong cases in this respect, ``(m) f+a'' selected an answer based on the random context (both for the contextual and the parametric answers), despite being unrelated to the question.

\subsection{Exposing Unseen Parametric Knowledge}
To understand the extent to which the parametric knowledge relies on pretraining, we count the percentage of parametric answers that were not seen as answers to other questions during fine-tuning. We use the counterfactual test set. 
For ``(m) f+a'', 2
5\% of the answers were not seen in the training data. For ``(m) f+cf'' this is the case for 26\%  of the answers, but most of them are identical to the contextual answer. 
For the ``(s) cb'' model, 23\% of the answers were not seen during fine-tuning. Finally, for the ``(m) f+cf+a'' 18\% were not seen, with disentangled answers 85\% of the times. 
We manually inspect those unseen answers, finding that some of them are correct with respect to world-knowledge although they contradict the context, as seen in Figure~\ref{fig:intro_example} and Table~\ref{tab:examples_con-par}. Overall, we see that while the models extract parametric answers from the pretraining, they have a strong tendency to repeat answers from fine-tuning.

\section{Related Work}

\paragraph{Knowledge Memorization.}
 
Language models are known to store factual knowledge memorized during pretraining. \citet{petroni-etal-2019-language} used ``fill-in-the-blank'' cloze statements to recover internal factual knowledge. \citet{roberts-etal-2020-much} trained QA models in a closed-book manner, without access to any external context. \citet{lewis-etal-2021-question} studied the overlap between the training and development sets of open domain benchmarks, including NQ, and showed that all models suffer from this issue, and perform worse on questions that do not overlap in their answers with the training data.
 \citet{dhingra-etal-2022-time} proposed to improve the memorization of versions of knowledge across time in language models, by adding a timestamp prefix in the pretraining input. They experimented with closed-book QA to evaluate the model memorization. 
\citet{akyurek2022tracing} focused on tracing the training examples that provided evidence for recalled facts from LMs, \citet{zhu2020modifying} tried to make transformers forget specific old facts and explicitly memorize new ones, 
while \citet{dai-etal-2022-knowledge} and \citet{https://doi.org/10.48550/arxiv.2202.05262} studied neurons and neuron activations that are associated with specific facts.


\vspace{-3px}

\paragraph{Knowledge Conflicts.}
\citet{longpre-etal-2021-entity} defined knowledge conflicts as cases where the contextual information contradicts the memorized information. To simulate this, they substitute entities in the gold context with another entity, showing over-reliance on the memorized knowledge. They suggested mitigating these conflicts by augmenting the training data with substituted instances. 
Other works addressed outdated facts or incorrectly induced pieces of information. For example, \citet{verga-etal-2021-adaptable} and \citet{de-cao-etal-2021-editing} created methods for modifying  unexpected parametric knowledge or incorporating newly
injected facts without the need for retraining or fine-tuning.


\paragraph{Answerabilty.}

SQuAD 2.0 \citep{rajpurkar-etal-2018-know} added unanswerable questions to SQuAD \citep{rajpurkar-etal-2016-squad}, providing a useful resource for identifying unanswerable cases in extractive QA systems. 
\citet{yatskar-2019-qualitative} found that the unanswerable questions in SQuAD 2.0 mostly represent cases of ``extreme confusion'' and are thus easy to detect. \citet{sulem-etal-2021-know-dont} extended SQuAD 2.0 by adding more challenging unanswerable examples.
\citet{asai-choi-2021-challenges} identified answerabilty as one of the two main challenges in information-seeking queries.
\citet{kim-etal-2021-linguist} focused on a subset of NQ questions that contain failed presuppositions, and are therefore unanswerable. This subset does not overlap with our data.


To the best of our knowledge, our work is the first to propose multiple answers, counterfactual augmentation and answerability augmentation to encourage and evaluate disentanglement, and to show that those approaches are complementary.
 
\textit{A concurrent work by \citet{li2022} has explored similar ideas and was brought to our attention while working on the manuscript.} 

\section{Conclusion}


We proposed a new method for disentangling and controlling whether the output of a LM should rely on its parametric knowledge or a given context. The method is simple and can be straightforwardly applied to a variety of LM architectures. We presented an extensive empirical evaluation and analysis of the method using different data augmentation approaches, showing that they are essential and complementary in allowing proper disentanglement, with improved robustness on counterfactual examples and an improved ability to deem questions unanswerable. In future work, we would like to extend this approach to the pretraining stage of LMs to allow even better disentanglement from the get-go. We hope this work will encourage more progress on models that disentangle parametric and contextual knowledge, towards more trustworthy and useful technology.


\section*{Ethics Statement}\label{sec:ethical}
We do not find any ethical considerations stemming from this work. Quite the contrary, we believe that disentangling knowledge sources to encourage the statements that an LM generates to be attributable \cite{rashkin2021measuring} can have a positive effect on the ability to avoid unwanted artifacts (that may otherwise be toxic or harmful).


\section*{Acknowledgements}
This work was carried out as part of a Master Sponsored Research Agreement between the Hebrew University and Google, and was also supported by a gift from Google. We thank Google Cloud for providing us with credits for running experiments on the Google Cloud Platform.

\bibliographystyle{acl_natbib}
\bibliography{anthology,custom}

\newpage
\appendix

\section{Limitations}\label{sec:limitations}
We discuss the following limitations of our work. First, the counterfactual data augmentation procedure we used can only be employed for questions whose answers are named entities. This restricts the applicability of the method as knowledge conflicts can arise for other types of questions, such as Boolean questions \citep{clark-etal-2019-boolq}. Extending our framework to other question types will require a new counterfactual data augmentation method. 

Second, we conduct our experiments using gold passages -- i.e., an oracle retriever. Using retrieved passages, which is often required in real-world applications, may introduce additional challenges when considering knowledge disentanglement.
Furthermore, the answerabilty approach presented in section \ref{sec:answerability} mainly serves as a proof-of-concept. It is quite simplistic, because the random context is unrelated to the question in terms of topic and participating entities. Future creation of unanswerable examples would include more distracting contexts, that at first glance seem very relevant, but still do not contain the answer.

We note another minor limitation, implied by the high accuracy in the counterfactual case relative to the factual accuracy (see \S\ref{subsec:results_parametric}). This might stem from the model's ability to identify that the text in the counterfactual examples is somewhat unnatural. It is therefore an indication of a potential limitation of the data augmentation methodology, albeit not a major one, judging by the small magnitude of the differences between the counterfactual and factual examples.

Finally, while our results indicate that models can learn to disentangle contextual and parametric knowledge, it remains unclear what characterizes easy vs. difficult cases for disentanglement. One such attribute, for example, can be the frequency of a given fact in the pretraining data. We view this as an important research question, which we plan to address in future work.

Due to the size of the models, we do not perform multiple trials of training from different initializations to test for significance. However, we do find similar trends across model sizes, which lends further support to the results presented.


\section{Technical Details}\label{ap:sec:setup}
We use the T5X library \cite{roberts2022scaling}. For inference we perform greedy decoding of the answers.
We trained for 50k training steps with constant learning rate of 0.0001 with a batch size of 32. We select the best checkpoint on the \emph{factual} validation set, prioritizing the standard performance criteria for QA models. The model sizes are 770M for T5-large and 11B for T5-11B. Each XXL training was done on 10 TPU hours. We did not try other hyperparameters.

\section{Additional Results} \label{sec:additional_results}
The following tables show results for the T5 large model (Tables
 \ref{tab:large_results_parametric},
 \ref{tab:large_similarity},
 \ref{tab:large_contextual}, \ref{tab:large_results_answerabilty}), and results on examples excluding context that contains only tables and not text (Tables \ref{tab:no_tables_contextual}, \ref{tab:no_tables_parametric}). We further report the accuracy on the no answer overlap development set (Table~\ref{tab:nao_param}) .
\label{ap:sec:results}
\begin{table}[ht]
\resizebox{0.98\columnwidth}{!}{%
\centering
\begin{tabular}{|c|c|c|c|c|}
\hline
{} &   Factual &  Counterfactual &     Empty &    Random \\\hline
(s) cb    &   - &          - & 10.26 &   - \\\hline
(m) f+cf  &   63.66 &          12.97 &  7.03 &   3.96 \\\hline
(m) f+a   &   77.14 &           2.86 & 14.43 &  12.01 \\\hline
(m) f+cf+a &   72.82 &          22.34 & 16.34 &  16.92 \\\hline
\end{tabular}}
\caption{Accuracy (in percent) of the parametric answer for the T5-Large models.}
\label{tab:large_results_parametric}
\end{table}

\begin{tabular}{lrrrr}
\toprule

\bottomrule
\end{tabular}

\begin{table}[ht]
\resizebox{0.98\columnwidth}{!}{%
\centering
\begin{tabular}{|c|c|c|c|c|}
\hline
{} &  Factual &  Counterfactual &  Empty &  Random \\\hline
\midrule
(m) f+cf  &   79.19 &          57.22 & 95.46 &  83.66 \\\hline
(m) f+a   &   99.78 &          99.71 &  0.00 &  35.82 \\\hline
(m) f+cf+a &   93.85 &          33.99 &  0.00 &   1.03 \\\hline
\end{tabular}}
\caption{Answer Separation: similarity between the contextual and parametric answers on the T5-Large models (in percent).}
\label{tab:large_similarity}
\end{table}

\begin{table}[ht]
\centering
\begin{tabular}{|c|c|c|}
\hline
{} &  Factual &  Counterfactual \\ \hline
\hline
(s) f (vanilla) &   76.34 &          67.84 \\ \hline
(s) f+cf     &   75.75 &          76.04 \\ \hline
(m) f+cf    &   76.12 &          77.73 \\ \hline
(m) f+a     &   77.14 &          66.37 \\ \hline
(m) f+cf+a   &   74.87 &          81.03 \\ \hline
\end{tabular}
\caption{Accuracy of the contextual answers for the T5-Large models (in percent).}
\label{tab:large_contextual}
\vspace{-15pt}

\end{table}

\begin{table}[ht]
\centering
\begin{tabular}{|c|c|c|}
\hline
{} &   Empty &  Random \\ \hline
\hline
(m) f+a   & 100.00 &  63.81 \\ \hline
(m) f+cf+a & 100.00 &  98.61 \\ \hline
\end{tabular}
\caption{Answerabilty scores for the T5-Large models (in percent).}
\label{tab:large_results_answerabilty}
\vspace{-15pt}

\end{table}

\begin{table}[ht]
\centering

\begin{tabular}{|l|c|c|}
\hline
{} &  Factual &  Counterfactual \\
\hline
(s) f (vanilla) &   86.79 &          79.23 \\\hline
(s) f+cf     &   88.10 &          91.43 \\\hline
(s) f+cf+a    &   87.50 &          95.77 \\\hline
(m) f+cf    &   87.70 &          89.82 \\\hline
(m) f+a     &   87.30 &          79.03 \\\hline
(m) f+cf+a   &   86.19 &          96.37 \\\hline
\end{tabular}
\caption{Accuracy for contextual answer on the test set without tabular contexts (73\% of the data did not include tables)}
\label{tab:no_tables_contextual}
\vspace{-15pt}

\end{table}

\begin{table}[ht]
\centering
\small
\begin{tabular}{|l|c|c|c|c|}

\hline
{} &  Factual &  Counterfactual &  Empty &  Random \\
\hline
(s) cb    &   - &          - & 25.40 &  - \\\hline
(m) f+cf  &   87.70 &           6.65 & 17.34 &  13.91 \\\hline
(m) f+a   &   87.30 &           0.71 & 22.78 &  23.89 \\\hline
(m) f+cf+a &   81.96 &          44.86 & 28.53 &  30.95 \\\hline
\end{tabular}
\caption{Accuracy for parametric answer on the test set without tabular contexts (73\% of the data did not include tables)}
\label{tab:no_tables_parametric}

\end{table}

\begin{table}[ht]
\centering
\resizebox{0.98\columnwidth}{!}{%
\begin{tabular}{|l|c|c|}
\hline
{} & Factual $\uparrow$ (diff $\downarrow$) & Counterfactual $\uparrow$   (diff $\downarrow$) \\ \hline
(s) f (vanilla)       &  78.11  (1.23) &         69.82  (-3.01) \\ \hline
(s) f+cf           &  79.88  (0.85) &         82.25  (-2.62) \\ \hline
(s) f+cf+a          &  76.63  (1.69) &         86.98 (-2.00) \\ \hline
(m) f+cf          &  77.51  (2.86) &         79.59  (-2.67) \\ \hline
(m) f+a           &  78.99  (1.23) &         70.12 (-5.5) \\ \hline
(m) f+cf+a         &  74.85  (3.25) &         87.28  (-2.37) \\ \hline
\end{tabular}}
\caption{Contextual Answer accuracy predicted on the No Answer Overlap (NAO) Dev set. In brackets, difference from total accuracy reported on the Dev set (Answer overlap + No Answer Overlap).}
\label{tab:nao}
\vspace{-15pt}
\end{table}

\end{document}